**RESEARCH ARTICLE**

# Research on the Inverse Kinematics Prediction of a Soft Biomimetic Actuator via BP Neural Network

HUICHEN MA[1], (Graduate Student Member, IEEE), JUNJIE ZHOU[2], JIAN ZHANG[1], AND LINGYU ZHANG[1,2]
[1]School of Mechanical Engineering, Beijing Institute of Technology, Beijing 100081, China
[2]Institute of Advanced Technology, Beijing Institute of Technology, Jinan 250300, China

Corresponding author: Junjie Zhou (bit_zhou50082@163.com)

This work was supported by the National Key Laboratory of Vehicular Transmission of China under Grant JCKYS2019208005.

**ABSTRACT** In this work, we address the inverse kinetics problem of motion planning of soft biomimetic actuators driven by three chambers. Soft biomimetic actuators have been applied in many applications owing to their intrinsic softness. Although a mathematical model can be derived to describe the inverse dynamics of this actuator, it is still not accurate to capture the nonlinearity and uncertainty of the material and the system. Besides, such a complex model is time-consuming, so it is not easy to apply in the real-time control unit. Therefore, developing a model-free approach in this area could be a new idea. To overcome these intrinsic problems, we propose a back-propagation (BP) neural network learning the inverse kinetics of the soft biomimetic actuator moving in three-dimensional space. After training with sample data, the BP neural network model can represent the relation between the manipulator tip position and the pressure applied to the chambers. The proposed algorithm is more precise than the analytical model. The results show that a desired terminal position can be achieved with a degree of accuracy of 2.46% relative average error with respect to the total actuator length.

**INDEX TERMS** Soft robotics, neural network, predictive models, bending, mechanical modeling, experiment.

## I. INTRODUCTION

In recent years, there have been expanding interests and increasing advancements in soft robots. Compared with traditional rigid robots, soft robots have irreplaceable advantages due to their intrinsic softness, such as high flexibility, good environmental adaptability, safe interaction with the surroundings, etc. [1], [2]. They play an irreplaceable role in the development of society. With the development of new material technologies, many materials have been studied and applied to soft robots [3], such as soft silicone rubber material [4], shape memory alloy (SMA) [5], and dielectric elastomer (DE) [6]. Therefore, soft robots could make a significant impact in many areas, including industrial applications [7], medical rehabilitation training devices [8], and bionic robots [9], [10].

One class of soft robot systems is soft biomimetic actuators (SBAs) with internal fluidic channels [11]. They are moved by the fluid and made of hyperelastic materials, which deform upon the pressurization of the internal channels to generate moving [12], [13]. The motion response of SBAs is governed by their morphology, which is defined by the geometry of the internal fluidic channels and the material properties used in fabrication. During the fabrication of SBAs, carbon fiber reinforced plastic is added to constrain the motion posture and significantly improve their strength, resulting in bending motion similar to human fingers and arms [14]. Therefore, as SBAs are used as soft biomimetic robotic arms and soft biomimetic robotic fingers, different structure designs and drive pressures will produce multistate 2D and 3D movements [15], [16].

The associate editor coordinating the review of this manuscript and approving it for publication was Jesus Felez.

  

The main challenge in building a soft biomimetic actuator is the control of such highly flexible structures. Compared with rigid robotic actuators, it is difficult to accurately model and control the forward and inverse kinematics of SBAs due to both material and geometric nonlinearities. For SBAs, the solution to the inverse kinematics problem is essential to generate paths in the task space to perform grasping or other tasks. It is a challenging task to solve the inverse kinematics. As for the research on inverse kinematics modeling and characterization of SBAs, previous efforts can be classified into two groups: 1) model methods and 2) model-free methods.

Model methods follow analytical or numerical approaches [17]–[19]. When applying a model method, the tradeoff between computing time and accuracy needs to be made [20]. For continuum SBAs, like for rigid robots, we can differentiate the direct kinematics model to find the manipulator tip position kinematics model, i.e., the linear transformation of the tip position velocity into the actuation variables velocity. Subsequently, in order to improve the prediction accuracy of the model, the kinematics of soft robots are mainly derived based on the piecewise constant curvature approximation [21]. Furthermore, many dynamics used either Euler/Lagrange or Lagrange formulations with quasi-static simplification [22]–[24]. The linear transformation is realized by the Jacobian matrix. These models may not accurately capture the complex nonlinear dynamics of soft robots. In part, specific approximations and assumptions of shape are required, and the computational hardness of the Jacobian matrix needs to be solved [25]. In the second place, an accurate material property model is needed. When SBAs are made of different materials, designed for complex structures, or added embedded components, the theoretical model will become more complex. A more complex model may solve these problems, but the process of deriving such models is tedious and brings some difficulty to the calculation [26]–[28]. Thus, due to the assumptions in the reduced model, the progressive computation accumulates numerical errors along with time steps, which brings in the accuracy problem for the SBA deformation.

The aforementioned challenges can be tackled via model-free methods based on machine learning approaches. The machine-learning and computer vision are mainly applied to model-free methods. Initial research into combining soft robotics with machine learning approaches can be traced back to Elgeneidy et al. [29], who used a purely data-driven approach for modeling the bending of soft pneumatic actuators. In addition, Van Meerbeek et al. [30] proposed various machine learning techniques to predict the type and magnitude of deformation in terms of twisting and bending of soft optoelectronic sensory foams with proprioception. Bruder et al. [31] identify the dynamical models of a soft arm using the Koopman operator theory for trajectory tracking via model predictive control. Apart from that, only a few researchers consider using neural networks (NNs) to learn inverse kinematics. In traditional robotics, neural networks (NNs) have been widely employed for learning inverse kinematics. In soft robotics, a long short-term memory (LSTM) network and feed-forward neural (FFN) network are used as the kinematic and force model [11], [12], [32], [33]. Nonetheless, these works did not address the issue of inverse dynamics of soft actuators, that is, the ability to estimate the input signal according to the moving state of actuator systems.

To further analyze and solve the inverse kinematics problem of SBAs, this paper combines the soft biomimetic actuator and BP neural network algorithm. The main contributions of this work are summarized as follows.

1) The BP neural network method is introduced into modeling the inverse dynamics of the soft biomimetic actuators. Although the BP neural network method has long been known, there have been very few attempts in this area.

2) Considering the difficulty of the soft actuators modeling: the relation between the stress and the strain is nonlinear, and the large deformation of the silicone material is complex. The accuracy of the model-based approach for inverse dynamics for soft actuators is limited, then the model-free approach based on the BP network could be more suitable for this particular case.

3) The present work details a parameter optimization procedure of the BP network. Not only the implement but also the experimental validation is proposed for this approach. Results show that the accuracy of the BP model-free approach is higher than a traditional model. This work provides a new direction for modeling the inverse dynamics of future soft actuators.

## II. ACTUATOR DESIGN AND FABRICATION

Fig. 1(a) shows the main structures of the SBA: 1) High elastic silicone driver matrix; 2) Three fiber-reinforced chambers; 3) Pneumatic connectors; 4) High hardness silicone molded sealed cap; 5) A metal base connector; 6) Pneumatic hoses.

The soft actuator is composed entirely of soft materials. The three elastic chambers dispose at 120° apart, and three smaller internal passages are disposed at 120° apart for weight loss. The radial restraint of the three chambers takes the form of left-right symmetrical double spiral fiber winding in this design, and the winding angle of the fiber is $\pm 3°$. In this way, the elastic fiber provides tension along the direction to exert large circumferential stress. When filled with fluid, the elastic air chamber only extends axially. The three chambers connect to different air valves that provide pressurized fluid through a pneumatic hose inserted at the top of each chamber.

Afterward, the actuator deforms under the action of three trigonally symmetrical pressurized chambers. When the pressures in the chambers are equal, the chambers elongate to the same length, leading to an axial stretching of the whole actuator. When the pressures in the chambers are not equal, the lengths of the chambers differ from each other, which allows the soft actuator to bend in any direction besides stretching in an axial direction. The metal base connector can be used to match the rigid mounting plate on the experimental platform to ensure the stability of the installation. Meanwhile, the connector is designed for expansibility. It can be further stuck to another soft actuator or other modules.

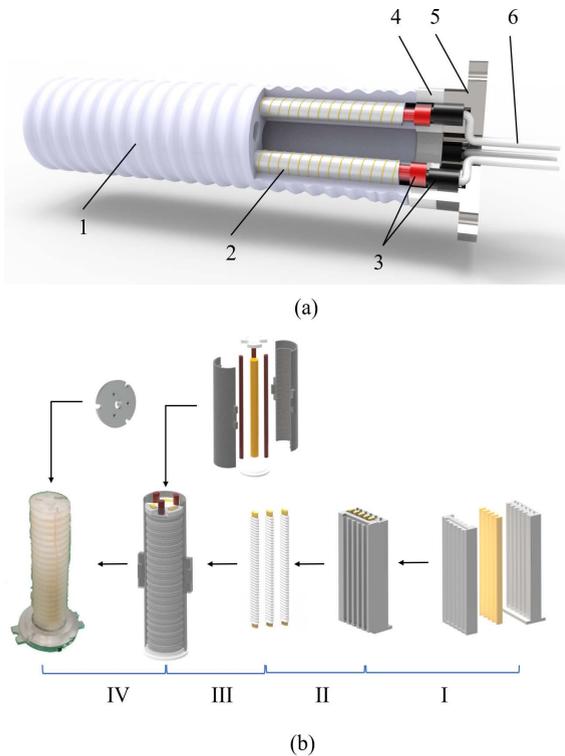

(a)

(b)

**FIGURE 1.** Schematic of: (a) the soft biomimetic actuator: 1 - High elastic silicone driver matrix, 2 - Three fiber-reinforced chambers, 3 - Pneumatic connectors, 4 - High hardness silicone molded sealed cap, 5 - A metal base connector, 6 - Pneumatic hoses; (b) the fabrication procedure of the soft biomimetic actuator: I - The mold pressing of the elastic chamber, II - The fiber winding, III - Actuator matrix molding, IV - The actuator end seal encapsulation.

The procedure involved in fabricating the soft actuator can be summarised in the following steps, as shown in Fig. 1(b):

I: The mold pressing of the elastic chamber. First, the chamber core and chamber casting mold are 3D printed with ABS material, and the inner layer elastic air chamber with a thickness of 1.5mm is cast with low-hardness silicone. The silicone consists of parts A and B (1 A: 1B by weight). Before assembling the molds, the release agent (LW-366, LONG WEI, China) is used to treat the surface of the molds to reduce the adhesion between silicone and ABS molds.

II: The fiber winding. Kevlar fibers are used to envelop the elastic chambers and restrain their radial expansion.

III: Actuator matrix molding. Insert the fiber-reinforced elastic chambers into the positioning hole of the ABS base plate, and cast the main part of the soft actuator with low hardness silicone. After curing, the joint sealed cap part is cast with high-hardness silicone.

IV: The actuator end seal encapsulation. Fix the metal base connector with a silicone adhesive to the actuator matrix after stripping, and finally, connect the joint with pneumatic hoses.

## III. METHODS
BP neural network is a kind of multilayer feed-forward with forward information propagation and error back-propagation.

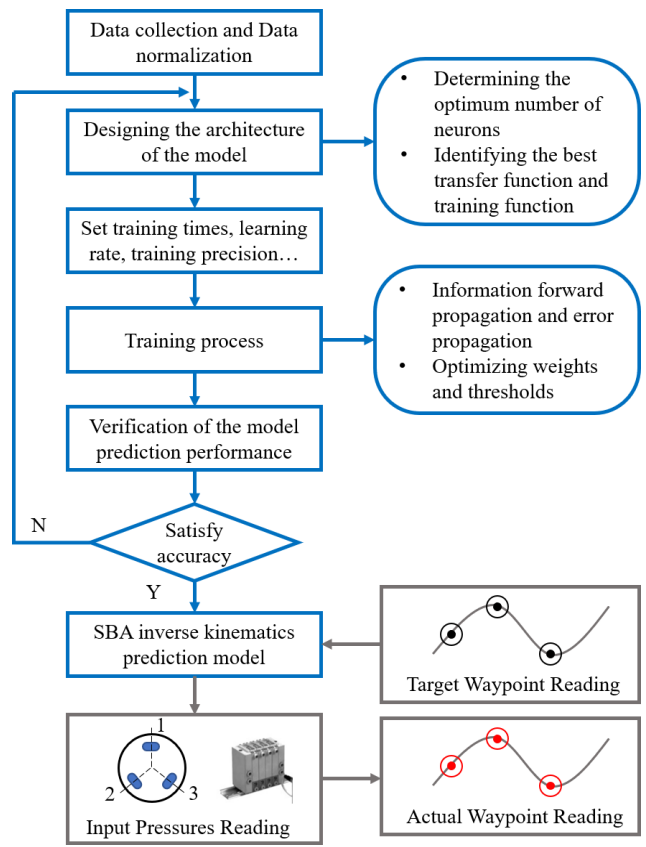

**FIGURE 2.** The developing procedure of the BP neural network model.

Compared with the traditional curvature approximation method, the BP neural network method is more suitable for processing nonlinear and complex system problems, preferably due to its complex self-learning and adaptive capabilities, which can greatly increase the fitting accuracy. And it does not need to filter and denoise the experimental data. Thus, we propose a BP neural network architecture based on finite experimental data of tip coordinates of the soft actuator. This architecture could efficiently implement the prediction of the soft actuator motion. Fig. 2 illustrates the developing procedure of the BP model schematically. After simple data collection and data normalization from the SBA bending experiment, the data are used for the parameter optimization and training of the BP neural network model. Thus, the inverse kinematics algorithm from the manipulator tip position to the input pressure can be implemented. According to the planning path, we get input pressure signals at different waypoints in the planning path. Input these pressure signals and control the proportional valves to output driving pressures, the actuator autonomously achieves desired trajectories. And the high-precision target trajectory navigation and prediction are performed under BP network modeling.

### A. PLATFORM SETUP AND DATA SAMPLING
To explore the relationship between the pressure and manipulator tip position and validate the inverse dynamic modeling based on the BP network, we have set up an experimental platform for data sampling, as shown in Fig. 3.

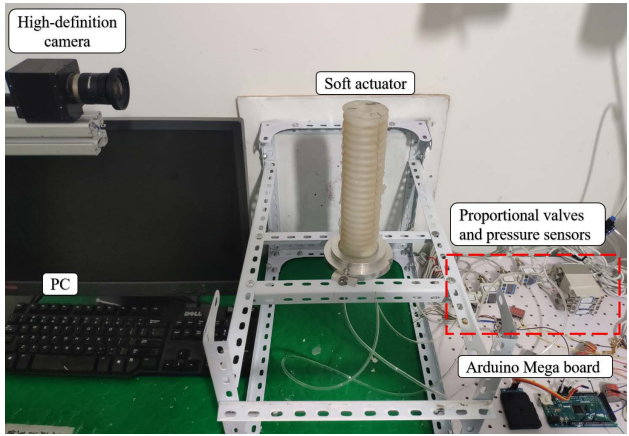

FIGURE 3. Overview of the soft biomimetic actuator experiments setup.

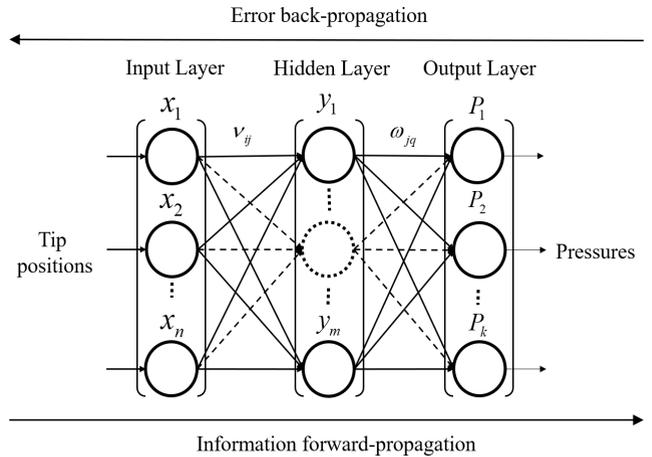

FIGURE 4. The structure diagram of a BP neural network [34].

The soft actuator is driven by a pneumatic driving device that can set the pressure of the three chambers via proportional valves (ITV0030-2BL, SMC, Japan). The control algorithm of proportional valves is programmed in Arduino. The timing of the actuation and the effective pressure supply are implemented through an Arduino Mega board (UNO R3, Arduino, Italy). Three pressure sensors (ISE80H-02-R, SMC, Japan) are used to collect the real-time pressures of the fluid in three chambers. Colored tape is attached to the ends of the soft actuator as a marker to facilitate coordinate recognition. Two high-definition cameras (MV-VD030SC, Microvision, China) are mounted on the two adjacent walls of the frame (500 mm×400 mm×400 mm. Two cameras are located 43 mm apart and used to collect the tip coordinates images of the soft actuator. Also, these images are conveyed through the PCI-E Ethernet interface. Finally, the coordinates of the mark points could be measured by a motion analysis software Tracker.

To demonstrate the value of the proposed kinematic model, we apply the input pressures to three chambers in various combinations of 0 kPa to 200 kPa at intervals of 40 kPa and test the tip positions of the soft actuator. It is important to notice that because the proportional valves are controlled by PWM waves, the pressures inputted into the chambers produce pulses. These pulses affect the accuracy of the data acquisition. For weakening the effect, five groups of experimental data are collected, and the average values are taken as the final experimental result. In total, 216 pairs of valid data (pressures and tip positions) are obtained. All data are also standardized to have zero mean and unit standard deviation. The data set is divided into the training set and the testing set. The data under the chamber 1 pressure of 0 MPa, 0.8 MPa, and 1.6 MPa serves as the training set. The data under the chamber 1 pressure of 0.4 MPa, 1.2 MPa, and 2.0 MPa serves as the testing set. The training set is used during the learning phase, whereas the testing set is only employed to evaluate the performance of the BP neural network model.

### B. PARAMETERS SELECTION OF THE BP NEURAL NETWORK

The BP neural network is a three-layer network that utilizes the connection weight to store information and fit functions. The layers contain the input layer, hidden layer, and output layer. And each layer has a series of nodes. The structure diagram of the BP neural network is presented in Fig. 4. The numbers of nodes in the input, hidden, and output layers of the BP neural network are defined as $n, m, k$, respectively. The connection weights are expressed as $v_{ij}$ and $\omega_{jq}$, which denote the weight from input node $i$ to hidden node $j$ and hidden node $j$ to output node $q$, respectively.

The number of nodes of the input layer is the number of the tip coordinates $(x, y, z)$, and therefore, the number of input nodes is 3. We predicted the input pressures $(P_1, P_2, P_3)$ of three chambers simultaneously, and therefore, the number of output nodes is 3.

The node number of the hidden layer affects the capacity of the BP network a lot: if the number of nodes is too small, it is less likely to produce the network with low precision; on the contrary, it is prone to oscillation and minimum local phenomenon may appear. In this paper, we use the empirical formula to calculate the number of hidden layer neurons in the training process, train and compare the different numbers of neurons, and thus derive the optimal neuron number of the hidden layer [35], [36].

$$N_{\text{hid}} = \sqrt{N_{\text{in}} + N_{\text{out}}} + \alpha \quad (1)$$

where $N_{\text{hid}}$ is the number of hidden layer neurons; $N_{\text{in}}$ and $N_{\text{out}}$ are the input node number and the output node number, respectively. $\alpha$ is a constant between 1-10.

Since the BP neural network model is a 3-input 3-output model, the number of hidden layer neurons ranges from 3 to 13. Even if the same number of hidden layer neurons are trained using the same data, the output results may be different for the uncertainty and complexity of the network. Therefore, the optimum neurons in the hidden layer of the network are determined by the method of trial and error in this study. Regression-coefficient ($R$) is frequently used to measure the differences between the values predicted by a model or an estimator. If a good model is expected, $R^2$ must be close to 1.

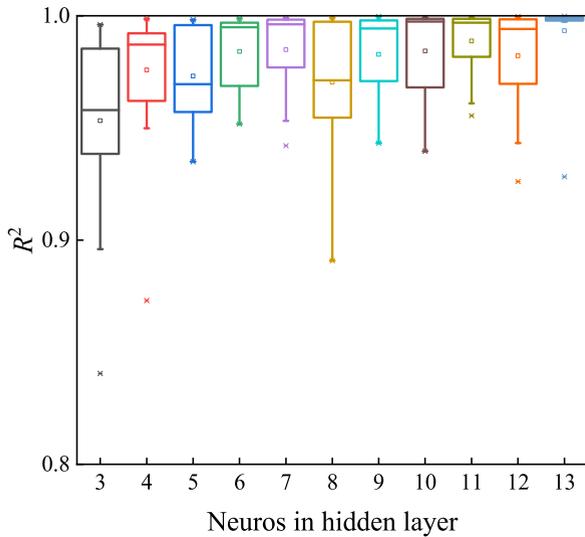

FIGURE 5. The $R^2$ values of different neuron numbers in the hidden layer.

Fig. 5 shows the training and verification results of a different number of neurons in the hidden layer, and it is found that the training error is the smallest when the hidden layer has 13 neurons. The mean value and median level under this number of neurons in the hidden layer are also the highest, representing that the network under this hidden neuron number is more stable. Then the network structure of the model is determined to be 3-13-3.

In order to avoid overfitting and falling into a local minimum, three parameters of the BP neural network structure are chosen: the training time $N_T$ that the algorithm is repeated, the training precision $N_P$, and the learning rate $\eta$ that determines the size of the step.

Training precision is set to 0.01 to meet the motion accuracy requirement of soft robots. The learning rate needs to be determined. Different learning rates can be selected for training, and the learning loss function results are shown in Fig. 6(a). The loss function doesn't improve when the learning rate is too low. The loss function begins to diverge if the learning rate is too high. Thus, the best learning rate is finally determined to be 0.01. Record its mean squared error after every ten iterations, and observe its convergence. As shown in Fig. 6(b), when the number of iterations is 500, the algorithm starts to converge, after which it remains stable, reaching the training precision value of 0.01. This shows that the algorithm model has a faster convergence speed. The number of iterations is 500.

Thus, the optimal parameters of the BP neural network are,
- $n = 3$
- $m = 13$
- $k = 3$
- $N_T = 500$
- $N_P = 0.01$

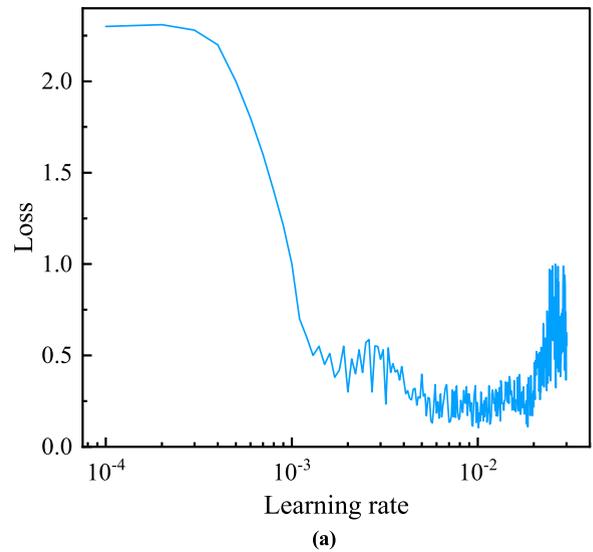
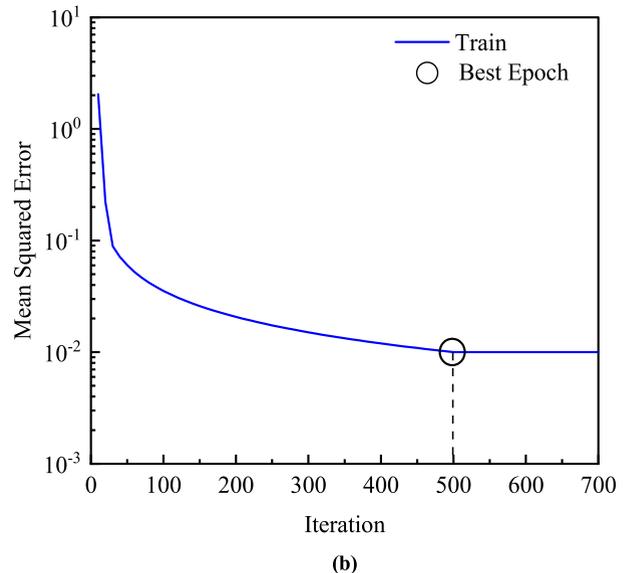

FIGURE 6. (a) Learning rate situation; (b) Best training performance is the iterations of 500 at the learning rate of 0.01.

- $\eta = 0.01$

### C. MODELLING BASED ON BP NEURAL NETWORKS

As shown in Fig. 4, in the error reverse propagation algorithm, the BP neural network continuously adjusts the weights and thresholds to approximate an arbitrary nonlinear function until obtaining a satisfactory output.

Assume that the input value is $x_i$, $i = 1, 2, \ldots, n$, the outputs of the hidden layer are calculated at first [37],

$$y_j = f_1 \left( \sum_{i=1}^{n} v_{ij} x_i + b_j \right) \quad (2)$$

where $j = 1, 2, \ldots, m$, $y_j$ denotes the output of hidden node $j$, $b_j$ represents the bias of hidden node $j$, and $f_1$ is the activation function of the hidden layer. In the current study, the sigmoid function is used for nonlinear relationships and represents an activation function for the respective neural layer. It has the advantages of being smooth, continuous, and differentiable and is more accurate than the linear function. In addition, the sigmoid function is not sensitive to the noise generated in learning, which can reflect the mainstream direction of a large number of data samples.

Then the output value is computed as follows,

$$P_q = f_2\left(\sum_{j=1}^{m}\omega_{jq}y_j + \beta_q\right) \quad (3)$$

where $p = 1, 2, \ldots, k$, $P_k$ represents the output of output node $k$, $\beta_q$ represents the bias of output node $k$, and $f_2$ is the activation function of the output layer.

The global prediction error $E$ is trained to achieve the minimum value via the back-propagation algorithm.

$$E = \frac{1}{2}\sum_{q=1}^{k}\left(P_q - R_q\right)^2 \quad (4)$$

where $R_q$ is the real output data and the expected output of the neural network.

Equation (3) is further expanded to the hidden layer.

$$E = \frac{1}{n}\sum_{q=1}^{k}\left[f_2\left(\sum_{j=1}^{m}\omega_{jq}y_j + \beta_q\right) - R_q\right]^2 \quad (5)$$

Equation (4) is then expanded to the input layer.

$$E = \frac{1}{n}\sum_{q=1}^{k}\left\{f_2\left[\sum_{j=1}^{m}\omega_{jq}f_1\left(\sum_{i=1}^{n}v_{ij}x_i + b_j\right) + \beta_q\right] - R_q\right\}^2 \quad (6)$$

Equation 5 is used in (3). $E$ is a function of parameters $v_{ij}, \omega_{jq}, b_j$, and $\beta_q$. Thus, the error $E$ will change as the values of these parameters change. During training, the weights and thresholds of the BP neural networks are modified to minimize the error $E$.

$$\begin{cases} \Delta v_{ij} = -\eta\frac{\partial E}{\partial v_{ij}} \\ b_j = -\eta\frac{\partial E}{\partial b_j} \\ \Delta\omega_{jq} = -\eta\frac{\partial E}{\partial \omega_{jq}} \\ \beta_q = -\eta\frac{\partial E}{\partial \beta_q} \end{cases} \quad (7)$$

where $\eta$ represents the learning rate.

After training, the nonlinear mapping from the tip coordinates $(x, y, z)$ of the soft actuator to the input pressures is established, that is,

$$(P_1, P_2, P_3) = f(x, y, z) \quad (8)$$

### D. MODEL EVALUATION ANALYSIS
The training set is used to train the BP neural network model with a 3–13–3 structure. At the same time, the testing set is used to verify the identification accuracy, as shown in Fig. 7. To evaluate the performance of the model, the mean

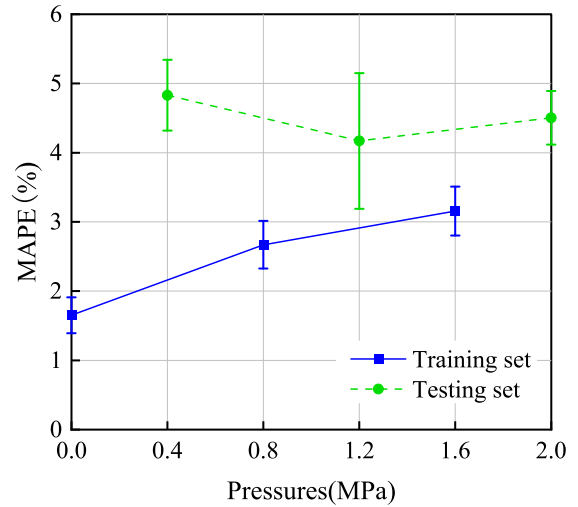

**FIGURE 7.** The MAPE of the BP neural network model.

absolute percentage error (MAPE) is applied. And this metric is defined as follows,

$$MAPE = \frac{1}{N}\sum_{i=1}^{N}\left|\frac{F_i - T_i}{T_i}\right| \times 100\% \quad (9)$$

where $N$ is the number of test data, $F_i$ is $i$-th forecast value, and $T_i$ is $i$-th actual value.

The results show that under different tip positions of the soft actuator, the maximum APE between the test value and prediction value is less than 6%. The MAPE for this model is less than 5%. The $R^2$ of the prediction model reaches 0.9767, which indicates that the predicted values are in good agreement with the actual values and could meet the prediction requirements of the inverse kinematics and the control demands of soft actuators.

The computational complexity calculation method for this proposed BP Network is obtained by the following equation [35]

$$O(b \times N_T \times (n \times m + m \times k)) \quad (10)$$

where $b$ is the number of training samples. The computational complexity of the above model is $O(88\,bt)$ by using (10). It can be seen that after determining the network structure, the computational complexity of this network is related to the number of samples $b$ trained and the number of iterations $N_T$. The richer training samples help reduce overfitting but also increase the computational complexity of training. In addition, the size of the number of training iterations $N_T$ has an impact on the computational complexity.

### IV. ANALYTICAL MODELLING
To verify our modeling approach, we performed the inverse kinematics analytical model [38]. This model conducts the open-loop control of the soft biomimetic actuator's tip position. The structural sketch of the soft biomimetic actuator under deformation is established in Fig. 8. The inverse kinematics process can be decomposed into three steps. The first step relates the tip coordinates $(x_0, y_0, z_0)$ of the SBA to the parameters $(l, \theta, \varphi)$. The second is the transformation from the parameters $(l, \theta, \varphi)$ to the lengths of chambers $(l_1, l_2, l_3)$. And the last one is from chamber lengths to the input pressures $(P_1, P_2, P_3)$.

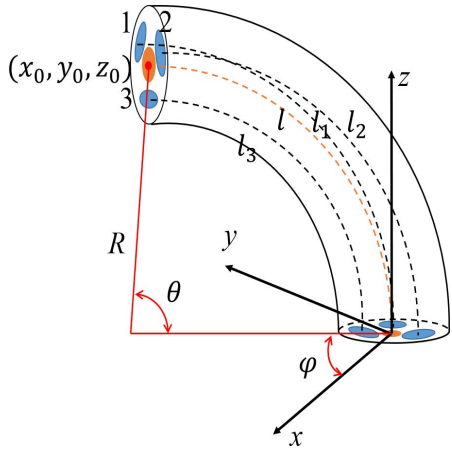

**FIGURE 8.** The structural sketch of the soft actuator under deformation.

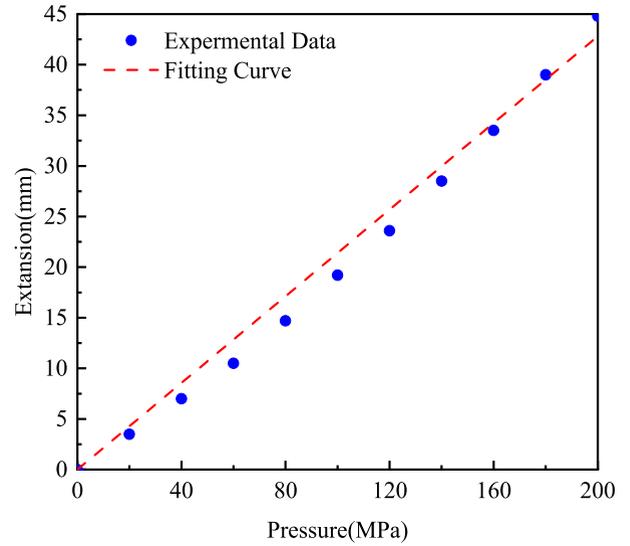

**FIGURE 9.** The actuator elongation for identifying the initial shear modulus of the SBA.

First, the azimuth angle $\varphi$, the bending angle $\theta$, and the axial length $l$ can be obtained from the tip position $(x_0, y_0, z_0)$ as

$$\varphi = \tan^{-1}\left(\frac{y_0}{x_0}\right) \quad (11)$$

$$\theta = \cos^{-1}\left(\frac{z_0^2 \cos^2\varphi - x_0^2}{z_0^2 \cos^2\varphi + x_0^2}\right) \quad (12)$$

$$l = \frac{z_0 \cdot \theta}{\sin\theta} \quad (13)$$

It should be noted that the bending angle $\theta$ would be miscalculated as zero when $x = 0$. So, in this case, we need (14) as an auxiliary.

$$\theta = \cos^{-1}\left(\frac{z_0^2 \sin^2\varphi - y_0^2}{z_0^2 \sin^2\varphi + y_0^2}\right) \quad (14)$$

Then, the axial lengths of three chambers $(l_1, l_2, l_3)$ can be represented with $(l, \theta, \varphi)$.

$$l_1 = l - \theta d \cos\left(\frac{\pi}{2} - \varphi\right) \quad (15)$$

$$l_2 = l - \theta d \cos\left(\frac{7\pi}{6} - \varphi\right) \quad (16)$$

$$l_3 = l - \theta d \cos\left(\frac{11\pi}{6} - \varphi\right) \quad (17)$$

where $d$ is the distance between the center of the soft actuator and the center of each chamber.

Based on the axial lengths of three chambers, the bending radius is given as

$$R = \frac{d(l_1 + l_2 + l_3)}{2\sqrt{(l_1^2 + l_2^2 + l_3^2 - l_1 l_2 - l_1 l_2 - l_1 l_3)}} \quad (18)$$

For pressures, we have the following description:

$$P_1 = \frac{l_1/l_0 - (l_0/l_1)^3}{k} \quad (19)$$

$$P_2 = \frac{l_2/l_0 - (l_0/l_2)^3}{k} \quad (20)$$

$$P_3 = \frac{l_3/l_0 - (l_0/l_3)^3}{k} \quad (21)$$

where $l_0$ is the internal length of the chamber, and $k$ is given as

$$k = \frac{A}{\mu_0 \cdot A'} \quad (22)$$

where $A$ is the stress area of the air chamber, $A'$ is the projected area of the actuator body material, and $\mu_0$ is the initial shear modulus of the silicone material.

## V. RESULTS AND DISCUSSIONS

In detail, the method process has been illustrated in Fig. 2. After simple data collection from the SBA bending experiment, the data are used for the neuron numbers in the hidden layer optimization and training of the BP neural network model. Once the parameters of the BP neural network are defined, the inverse kinematics algorithm from the manipulator tip position to the input pressure can be implemented.

To prove the practical value of the BP neural network model, we conduct the model-based open-loop control of the tip positions of the soft robotic arm and carry out a trajectory planning experiment. The target trajectory is an 8-shaped spatial curve at the projection onto the XOY plane. We choose 41 waypoints from it. According to the BP network prediction model and the analytical model in section IV, we can calculate the corresponding input pressure vectors with different methods.

In the analytical model, to identify the material parameters $\mu_0$ and $k$, three chambers are pressurized equally from 0 kPa to 200 kPa at intervals of 20 kPa, and the lengths are recognized. Fig. 9 presents the experimental data and the fitting curve. The parameter $k$ in (22) equals 2.128 MPa$^{-1}$, and the parameter $\mu$ is 1.197 MPa. According to the results, we can calculate the required pressures to the tip coordinates of the soft actuator through the inverse dynamics model.

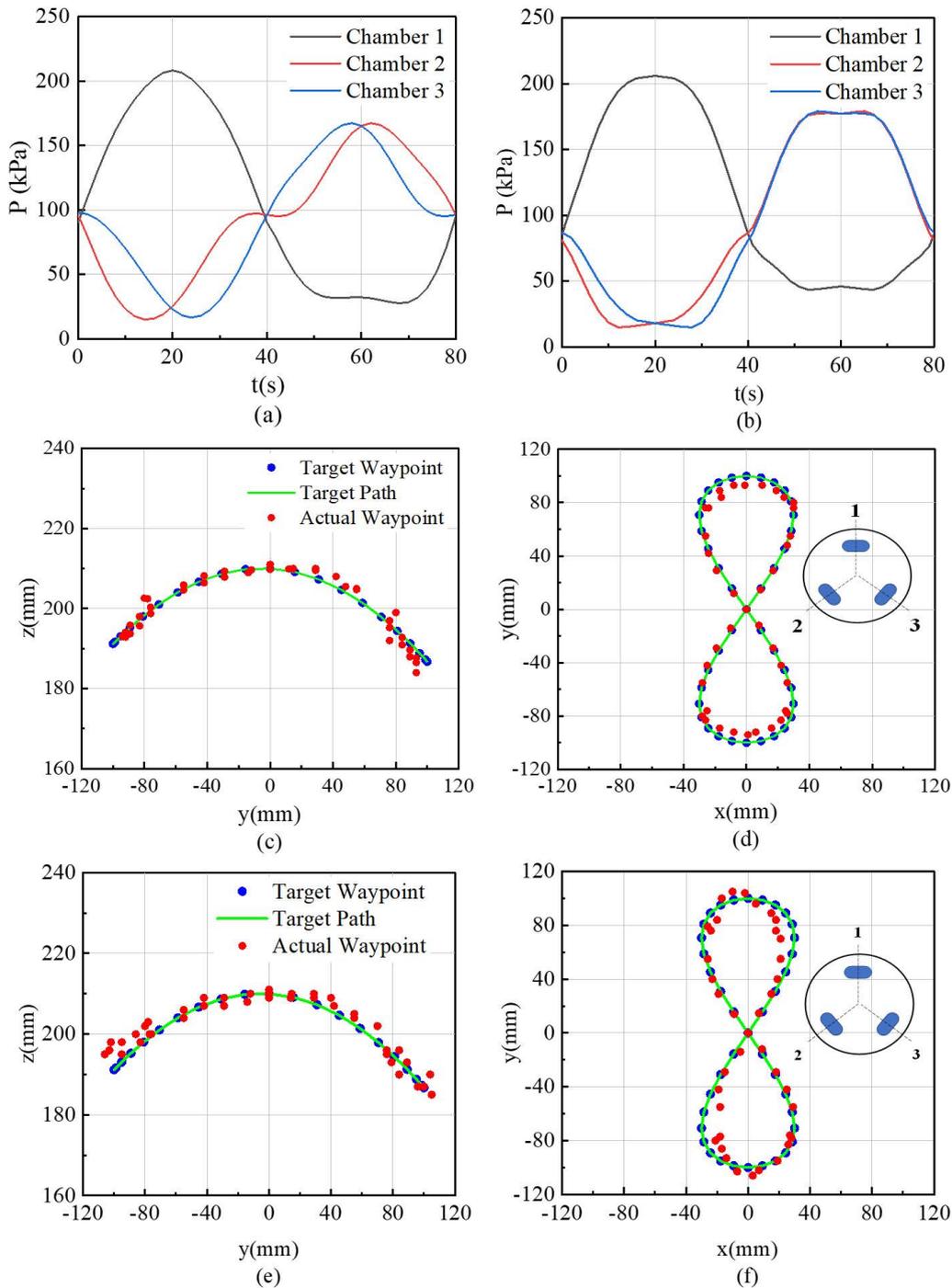

**FIGURE 10.** Pressure inputs to the three chambers and tip path of the soft biomimetic actuator: (a) Input pressures of BP network prediction model; (b) Input pressures of the analytical model. (c) Side view in the BP network model. (d) Top view in the BP network model. (e) Side view in the analytical model. (f) Top view in the analytical model.

In Fig. 10(a) and (b), the input signals of three-chamber pressure used to complete the trajectory are given. Fig. 10(c) to (f) shows the actual path at the tip of the SBA under input pressure vectors, the target route point selected to calculate the required pressure, and the actual route point in the top view and the side view, respectively. The details about measuring the coordinates of the mark points by the software Tracker are shown in the Supplementary Materials. The position error while following the path is shown in Fig. 11. For the 41 waypoints, the average position error, the standard deviation, and the maximum value are given in Table 1. The mean relative error relative to the total arm length is 2.46 %.

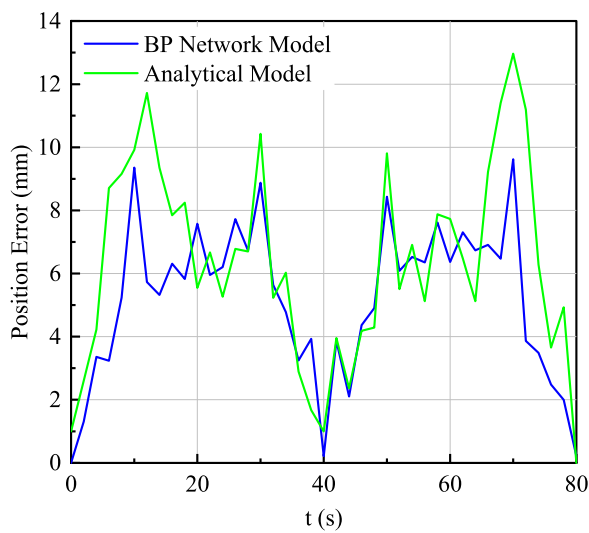

**FIGURE 11.** The experimental procedure followed in the model application section.

**TABLE 1.** Dimensions of experiment errors.

| Symbol | BP network model | Analytical model |
|---|---|---|
| Average position error[mm] | 5.17 | 6.34 |
| Maximum position error[mm] | 9.35 | 12.96 |
| Standard deviation | 0.38 | 0.49 |

Compared with the relative average error of 3.09% with respect to the total length of the soft actuator in the national analytical model, the experimental results demonstrate that the developed BP network model has more accuracy.

The performance of the model in the trajectory planning experiment is not perfect. The reasons for measurement error are analyzed. On the one hand, an open-loop static controller is not suitable enough for continuously moving control; on the other hand, the hysteresis of soft materials and the pulse shock signal of proportional valves lead to errors. On the other hand, since the proposed BP neural network scheme comprises training and validation, it requires a longer training time than the analytical method, which consists of only mathematical equations. Nevertheless, the longer training time here is justifiable as the proposed scheme is robust and more accurate.

Nevertheless, the results achieved in this work open a set of future investigations into the control of soft robots by implementing more advanced machine learning methods.

The experimental procedure is followed in the model application section.

## VI. CONCLUSION AND FUTURE WORK

The work presented here demonstrated an alternative approach for predicting and controlling the bending angle of a three-chamber pneumatic soft actuator using a model-free approach. We demonstrate the actuator with a modular structure and present its design and fabrication. Although some kinematic model studies have been carried out on the three-chamber soft actuator, it is difficult to get the exact solutions by a traditional analytical model. Few people combine the model-free methods with the kinematic model. Thus, this article puts forward a scheme aided by BP neural network modeling to realize nonlinear prediction and mapping from the tip coordinates ($x$, $y$, $z$) to the input pressures.

Thus, the conclusions can be summarized as follows:

1) Considering the difficulty of the modeling of soft actuators: the nonlinear relation between the stress and the strain and the complex deformation of the silicone material. The model-free method is introduced to solve the problems of high complexity and poor accuracy in the model method. Based on the BP neural network algorithm, a complete soft biomimetic actuator inverse kinematics framework that can predict the driving pressures of the SBA on curved trajectories is developed. By solving the model, we find the optimal parameters of the BP network computing framework for high-required precision.

2) The feasibility of the prediction model based on the BP neural network has been tested on the real soft biomimetic arm. The prediction model shows accurate and fast performance because the mean position error is 5.17 mm, the maximum position error is 9.35 mm, and the standard deviation is 0.38. The mean relative error relative to the total arm length is 2.46 %. All these error dimensions are less than those of a traditional analytical method.

The performance result of the model in the model application section is imperfect. The current problem is that this BP neural network model is an open-loop static controller, which is not suitable enough for continuous control. In addition, the manufacturing error and environmental change greatly affect the motion results of the soft biomimetic robotic arm. For future work, the combination of feedback control and damping is expected to achieve more precise control of the complex motion. Subsequently, the hybrid intelligence algorithm (e.g., combining BP neural network and genetic algorithms) is hopeful to further improve the accuracy of the inverse kinematics prediction of a soft biomimetic actuator. To explore the universality of this method, it will be validated with more complex soft actuators. The soft biomimetic robotic arm can be put to wide use in searching and rescuing in narrow workspaces, medical applications in long-distance surgical operations, and underwater operations.

### ACKNOWLEDGMENT
A preprint has previously been published [39].

### CONFLICT OF INTEREST
The authors declare that there is no conflict of interest.

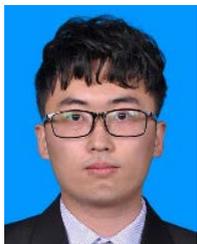

**HUICHEN MA** (Graduate Student Member, IEEE) was born in Hebei, Hengshui, China, in 1998. He received the B.S. degree from the College of Mechanical Engineering, Beijing Institute of Technology, Beijing, China, in 2020, where he is currently pursuing the Ph.D. degree with the School of Mechanical Engineering.

Since 2020, he has been working on soft robotics with the School of Mechanical Engineering, Beijing Institute of Technology. His research interests include soft robotics and applications, fluid power components, and systems.

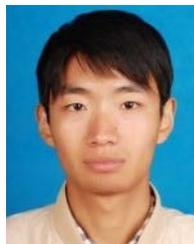

**JIAN ZHANG** was born in Shanxi, Datong, China, in 1994. He received the B.S. degree, in 2014, and the M.S. degree with the College of Energy and Power Engineering, Lanzhou University of Technology, Lanzhou, China, in 2017. He is currently pursuing the Ph.D. degree with the School of Mechanical Engineering, Beijing Institute of Technology, Beijing, China.

Since 2018, he has been working on soft robotics with the School of Mechanical Engineering, Beijing Institute of Technology. His research interests include soft robotics and applications, fluid power components, and systems.

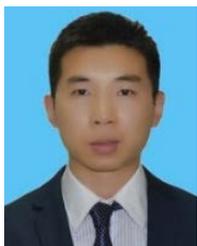

**JUNJIE ZHOU** was born in Hengshui, China, in 1986. He received the B.S. and Ph.D. degrees in mechanical engineering from the Beijing Institute of Technology, in 2009 and 2015, respectively.

From 2011 to 2013, he was a Visiting Scholar with the Maha Fluid Power Research Center, Purdue University. Since 2015, he has been an Assistant Professor with the Mechanical Engineering Department, Beijing Institute of Technology. He is the author of one book and more than 40 journals or conferences papers. His research interests include soft robotics and applications, fluid power components, and systems. He is a member of the Fluid Control Engineering Committee, Chinese Society of Mechanics.

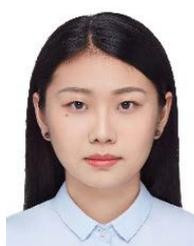

**LINGYU ZHANG** was born in Beijing, China, in 1996. She received the B.S. degree with the College of Mechanical Engineering, University of Science and Technology Beijing, Beijing, in 2019. She is currently pursuing the master's degree with the School of Mechanical Engineering, Beijing Institute of Technology, Beijing.

Since 2019, she has been working on soft robotics with the School of Mechanical Engineering, Beijing Institute of Technology. Her research interests include soft robotics and applications, especially on soft actuator components and systems.

• • •